\documentclass[10pt,twocolumn,letterpaper]{article}

\usepackage{epsfig}
\usepackage{graphicx}
\usepackage{abstract}
\usepackage{amsmath}
\usepackage{amssymb}
\usepackage{flushend}
\usepackage{nccmath}
\usepackage{xcolor}
\usepackage{color, colortbl}
\usepackage{float}
\usepackage[textsize=tiny]{todonotes}
\usepackage[backend=bibtex,
            style=numeric,
            natbib=true
            ]{biblatex}
\usepackage{hyperref}
\hypersetup{hypertex=true,
colorlinks=true,
linkcolor=black,
anchorcolor=black,
citecolor=blue}
\usepackage{indentfirst} 
\newcommand\method{HVDetFusion}
\definecolor{ForestGreen}{RGB}{34,139,34}
\definecolor{LightCyan}{rgb}{0.88,1,1}
\definecolor{Gray}{gray}{0.8}
\usepackage{geometry}
\title{HVDetFusion:  A Simple and Robust Camera-Radar Fusion Framework}

\author{Kai Lei, Zhan Chen, Shuman Jia, Xiaoteng Zhang\\
HascoVision Technology (Shanghai) Co.,\\
{\tt\small \{kail05, zhanc, shumanj, xiaotengz\}@hascovision.com}
}
\begin{document}
\date{July 15, 2023}

\twocolumn[
    \maketitle
    \begin{onecolabstract}
     In the field of autonomous driving, 3D object detection is a very important perception module. Although the current SOTA algorithm combines Camera and Lidar sensors, limited by the high price of Lidar, the current mainstream landing schemes are pure Camera sensors or Camera+Radar sensors. In this study, we propose a new detection algorithm called HVDetFusion,  which is a multi-modal detection algorithm that not only supports pure camera data as input for detection,  but also can perform fusion input of radar data and camera data. The camera stream does not depend on the input of Radar data, thus addressing the downside of previous methods. In the pure camera stream, we modify the framework of Bevdet4D for better perception and more efficient inference, and this stream has the whole 3D detection output. Further, to incorporate the benefits of Radar signals, we use the prior information of different object positions to filter the false positive information of the original radar data, according to the positioning information and radial velocity information recorded by the radar sensors to supplement  and fuse the BEV features generated by the original camera data,  and the effect is further improved in the process of fusion training. Finally, HVDetFusion achieves the new state-of-the-art 67.4\% NDS on the challenging nuScenes test set among all camera-radar 3D object detectors. The code is available at \href{https://github.com/HVXLab/HVDetFusion}{https://github.com/HVXLab/HVDetFusion}
    \end{onecolabstract}
]

\section{Introduction}
 3D detection technology is rapidly developing in the field of autonomous driving. The detection task based on the nuScenes dataset has also become one of the most popular competitions in the detection challenge task in recent years. Multiple sensors are used to collect different types of data in most autonomous driving scenes, such as a combination of cameras and range sensors as lidar and radar. How to effectively process information based on machine learning methods, how to integrate data from different sensors for perception training with accuracy improvement has also become one of the most important part of the work. BEVFormer\cite{li2022bevformer} use a transformer-based encoder to transfer the multi-camera inputs into BEV features. Bevdet\cite{huang2021bevdet} takes images from 6 cameras, it extracts image features and reconstructs fused image features from 6 cameras in BEV grid, and it encodes features and predicts the targets in bird’s-eye view(BEV), Bevdet4D\cite{huang2022bevdet4d} uses multiple sequence keyframes to improve the detection of Bevdet from 3D spatial domain to spatio-temporal domain with temporal information to access the temporal cues by querying and comparing the two candidate features which could greatly reduces the velocity error, PETRv2\cite{liu2022petrv2} utilizes the temporal information of previous frames to boost 3D object detection, BevDepth\cite{li2022bevdepth} targets on depth estimation on camera-based bird’s-eye view(BEV) 3D object detection. A camera-awareness depth estimation module is used to improve the depth predicting capability. CenterFusion\cite{nabati2021centerfusion} uses a center point detection network to detect objects by identifying their center points on the image. At the same time, it use a frustum-based method to associate the radar point cloud to corresponding image features. CRAFT\cite{kim2022craft} associates image proposal with radar points in the polar coordinate system, then use consecutive cross-attention based feature fusion layers to fuse camera and radar datasets.
 
Inspired by the design of many 3D detection models in autonomous driving, considering the flexibility and efficiency of the Bevdet model architecture, which can reasonably combine model design and actual needs under the condition of computing power constraints in actual scene, so we design a two-stage model structure based on the series of Bevdet model architecture, it supports a combination of multiple data types as input, which is called HVDetFusion. After training with camera\&radar data and continuously tuning the model structure, HVDetFusion achieves the new state of-the-art 67.4\% NDS on the challenging nuScenes test set among all camera-radar 3D object detectors. Different types of data processed in a decouplable branch structure, the plug-in branch corrects and optimizes the accuracy of the prediction content of the main detection structure. Which makes our model architecture support more scenarios, it is more convenient to select the appropriate model solution according to the actual situation under the premise of comprehensively considering the accuracy and running speed. The HVDetFusion structure will be specified in Section 3, and the experimental results will be specified in Section 4.

On this basis, through a large number of experiments, we found some effective tricks in the training and testing process of the model. These tricks also help our model to better capture and summarize the object information in the key frame, so that the training process is more effective, and these tricks will be specified in the ablation experiment part of this paper.

\section{Related Work}
\subsection{Vision-based 3D object detection}
3D object detection is the key perception task in autonomous driving. FCOS3D\cite{wang2021fcos3d} extends the 2D object detection problem to the 3D object detection problem, using the spatial correlation of the target in the camera image feature to detect the object and achieved good results, but the prediction accuracy needs further improvement in the targets' translation, velocity, and orientation. PGD\cite{wang2022probabilistic} presents geometric relation graphs to facilitate depth estimation for 3D object detection, it wants to improve the final result by using deep depth as a breakthrough point. Similarly, DD3D\cite{park2021pseudo} points out that depth pretraining can improve in 3D detection. BEVDepth\cite{li2022bevdepth} shows that auxiliary pixel-wise depth supervision improves the performance. How to effectively fuse multi-sensor data is also an important part in 3D detection tasks. Bevdet\cite{huang2021bevdet} extracts image features and the image features of multiple cameras are reconstructed and fused in the BEV grid, and encodes features in the BEV space, Bevdet4D\cite{huang2022bevdet4d} is the temporal extension of Bevdet\cite{huang2021bevdet}.
\subsection{Fusion-based Methods}
In order to make full use of existing data to detect 3D objects with higher precision, more sensors will be put into use, and how to effectively fuse camera images and range measurement information has become a very crucial and popular research topic. The more common idea is to attempt to improve depth estimation by projecting radar points to the image\cite{lin2020depth}. And\cite{long2021radar} learns a mapping from radar returns to pixels and a depth completion method is used subsequently. Centerfusion\cite{nabati2021centerfusion} uses a novel frustum-based method to associate the radar detections to points from camera images. Fusing the camera image and point feature maps under the BEV perspective is also a good choice, BEVFusion\cite{liu2022bevfusion} unifies multi-modal features in the shared bird's-eye view (BEV) representation space and fuse feature maps by element-wise concatenation, UVTR\cite{li2022unifying} preserves the voxel space without height compression to alleviate semantic ambiguity and enable spatial connections. the cross-modality interaction could get effective geometry-aware expressions in point clouds and context-rich features in images. CRN\cite{kim2023crn} transform perspective view image features to BEV with valid radar points. What’s more, it uses multi-modal deformable attention module to tackle the spatial misalignment. what’s more, TransCAR\cite{pang2023transcar} uses a sparse set of 3D object queries to index into 2D features from camera images, then applies transformer decoder to learn the interactions between radar features and vision-updated queries. CRAFT\cite{kim2022craft} process association between camera images and radar points in the polar coordinate system, then use consecutive cross-attention to fuse. 3D-CVF\cite{yoo20203d} uses a cross-view spatial feature fusion strategy. A gated feature fusion network is applied to blend functional and LiDAR features appropriately by region using a spatial attention map. CMT\cite{yan2023cross} encodes the 3D points into multi-modal features without view transformation, and directly outputs accurate 3D bounding boxes. Pai3d\cite{liu2022pai3d} extracts instance-level semantic information from images, then used to augment each LiDAR point in 3D detection network to improve detection performance. SparseFusion\cite{xie2023sparsefusion} transforms the camera candidates into the LiDAR coordinate space by disentangling the object representations. Then fuses the multi-modality candidates in 3D space with self-attention module. TransFusion\cite{bai2022transfusion} uses a soft-association mechanism to handle inferior image conditions. it's decoder uses a sparse set of object queries predicts initial bounding boxes from a LiDAR point cloud, and adaptively fuses the object queries with useful image features. DeepInteraction\cite{yang2022deepinteraction} use a multi-modal representational interaction encoder and a multi-modal predictive interaction decoder to maintain and learn individual per-modality representations. Benefit from the related work mentioned above, the point cloud processing branch in the HVDetFusion model fuses the feature-extracted image and the filtered point cloud data in the BEV space, which can improve the efficiency of fusion and remove the influence of redundant features.
\section{\method{} Method}
\begin{figure*}[h]
    \centering
    \includegraphics[width=16cm]{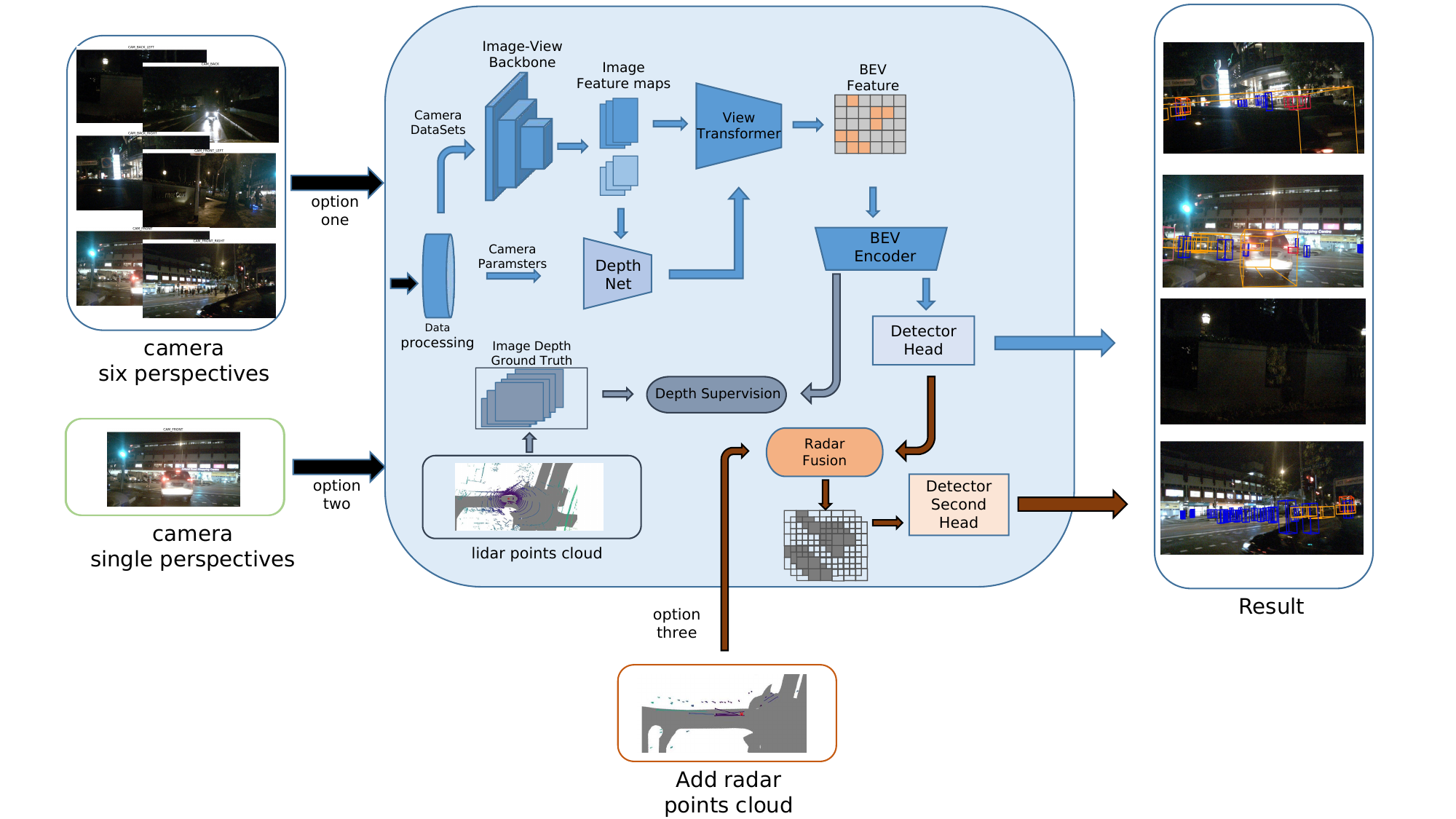}
    \caption{An image of clarify the process of radar data filtering and fusion}
    \label{fig:hvdet_process}
\end{figure*}
We designed a model structure based on Bevdet4D, which is called HVDetFusion. Based on the original architecture, we continued to optimize the fusion process of image features. At the same time, we designed decoupling fusion branch which is used to process radar data and extract radar point clouds. Effective spatial position information and radial velocity information assist the 3D detection modules to predict the process of target position, velocity, and direction, which improves the prediction accuracy of each component. The entire workflow is clarified in Figure \ref{fig:hvdet_process} including data acquisition, data processing, and feature fusion, target detection.

As shown in the figure, our model supports three different data combination methods as input. The input data of option 1 is the image data collected by cameras with 6 different viewing perspective, and the input data of option 2 is the camera using the front-view perspective. With the collected image data, the input data of option 3 refers to adding the radar point cloud on the basis of option 1 or option 2. In option 1 and option 2, we process the image data and perform data augmentation, and convert the image corresponding to the camera parameters. Feeding them to the image-backbone model structure to obtain the image feature map. The specified part of the image feature map are used as input of DepthNet, and finally another part of the image feature map and the output of DepthNet are transferred to the BEV grid through View-Transformer. At this point, the images of different viewing angles collected by the multi-camera have been fused. During the training process, the accuracy of the coordinates of the lidar point cloud is also used to supervise and adjust the BEV features, finally the predicted components are obtained through 3D Detection Head. In option 3, we filter the radar point cloud, and fuse the retained effective point cloud with the output features of the 3D Detection Head. The final feature obtained after fusion will obtain the corrected prediction component through the secondary detection head.
\subsection{Datasets Introduction}
In order to further improve the detection effect of the model, on the basis of the data collected by the camera, we fused the data collected by the radar sensors to supplement and correct the overall prediction results, so as to make up for the defect of depth information loss caused by the projection characteristics of the data collected by the camera sensor. We use the full nuscenes dataset which contained 1.4 million radar scans as a part of the fusion for training and testing. The radar scans comes from 5 radar sensors in the vehicle. The monitoring range of the sensor is within 200-300m, and the speed of the object is measured by the Doppler effect. However, the positioning of the radar sensor is more biased than that of the lidar sensor, and the recorded radar points are also more sparse\cite{caesar2020nuscenes}. In order to prevent false positive noise information from negatively affecting the training results of the model after fusion, it is necessary to filter the radar point cloud and retain the valid point cloud to participate in the subsequent fusion operation. It should be noted that the speed information \(v_x\) and \(v_y\)  contained in the radar point cloud are the components  of the instantaneous radial speed of the object in the current coordinate system, and there will be a certain difference between this variable and the actual speed of the vehicle\cite{nabati2021centerfusion}. In addition, the speed information recorded by the original radar sensors includes the original speed information and the compensated speed information. We use the compensated speed information as the effective value collected by the radar sensors to participate in the subsequent and fusion process.
\subsection{Radar Association}
We use Bevdet4D as the base structure of the detection task which uses images datasets collected by 6 cameras as input for training and prediction. In order to achieve the ideal fusion effect, on the original network architecture which is used to process pure camera datasets, we consider adding an auxiliary branch architecture for processing radar datasets which processes and filters radar data to obtain effective depth information and speed information, then performs feature fusion with the detection main branch. As mentioned above, radar point cloud could make up for the defect of the depth information in the process of collecting camera datasets. In order to achieve the ideal fusion training effect, in the stage of data processing, it is necessary to ensure that the position information of each object in the radar point cloud can correctly match the corresponding area of the image feature. On this basis, the position and velocity information recorded in the radar point cloud are fused with the image feature. Due to the characteristics of radar sensors, many objects in the scene will be recorded as part of the radar point cloud. These objects not only include the parts we are interested in, but also include other cluttered and invalid parts. Before fusion, we can use the distribution of image features to obtain the prior of the object position, use the position prior to filter the radar point cloud, and perform the next fusion operation on the point cloud whose position is successfully matched.

\begin{figure}[htp]
    \centering
    \includegraphics[width=7cm]{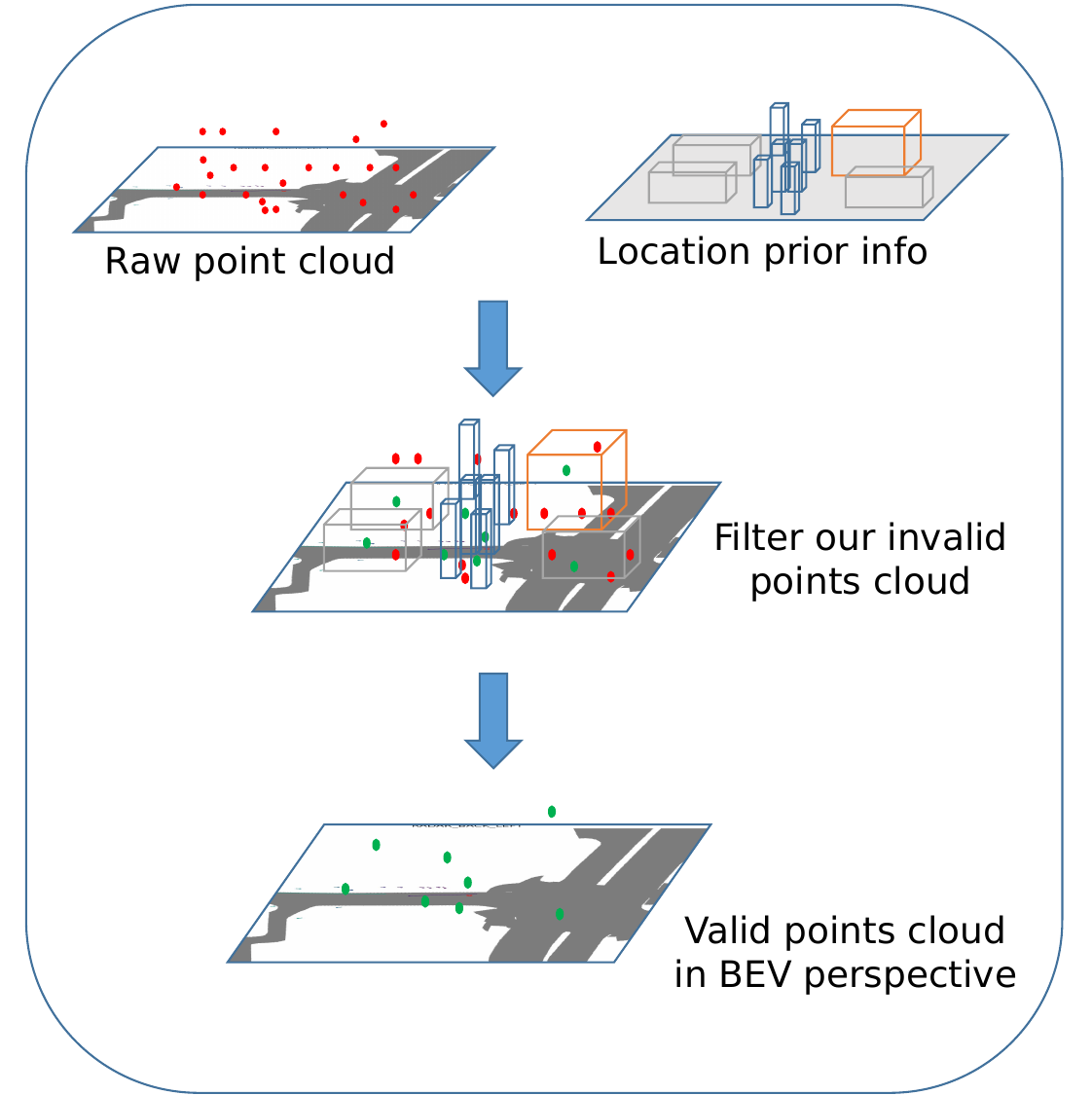}
    \caption{Demonstration of filtering radar points using the location prior information of the object.}
    \label{fig:radar_filter}
\end{figure}

    Figure \ref{fig:radar_filter}. shows the process of using the position prior information of the object in the BEV space to filter and obtain effective radar point cloud. The 3D boxes representing the position prior about objects of different categories, and different colors are used to distinguish the different categories. And we use red points to represent the original radar points, by the way green points is used to represent the matching points. The point cloud that is matched successfully will enter the next stage of matching and fusion process.

We process the point cloud position information into a 2D bounding box under the BEV (Bird's Eye View) perspective. In order to ensure that the radar points maintain a high position accuracy during the fusion process, if there is a situation where two bounding boxes are overlapped, regarding the midline position of the bounding box generated by the radar point projection as the new boundary of the overlapping area of the two bounding boxes. Then the position information of these bounding boxes corresponds to the projection position of the radar 3D point cloud under the BEV perspective. At the same time, we use the regression results of the position and size of objects under each category obtained through image features as position priors to generate bounding boxes ground truth in a certain sense, then calculating IOU scores between the 2D bounding box generated by radar point cloud and the ground truth on each key frame. Further, we preset the hyperparameter \(\alpha\) as the size scaling factor of the 2D bounding box, and the hyperparameter \(\beta\) as the threshold to control the difficulty of matching. By reducing  \(\beta\) or increasing \(\alpha\), each object area can contain more radar point cloud information. When the hyper parameters setting is reasonable, noise points with large position deviations can be filtered out. However, the proportion of effective radar data is increased in the fusion process at the same time. Both of these aspects are important to improve the robustness of fusion predictions.

After obtaining the filtered radar 2D bounding box, we process the position and velocity information of the point cloud corresponding to each 2D bounding box into a tensor of the radar feature map, and assign it to the corresponding area in the newly generated radar feature map, these areas correspond one-to-one to the bounding boxes that have been matched. The radar feature map is concatenate with the feature map calculated by BEV encoding modules, and the fused feature is used as the input of the second regression head to correct the deviation of the speed, rotation, and position regression of each category of tasks. The task of the second regression head is to further improve the accuracy of the predicted position, speed, and rotation angle predicted by the main regression head. During the training process, the fused radar feature will compare the speed and position contained in the original feature map. The information is effectively integrated, and the influence of redundant and biased information that still exists in the matching process of the radar point cloud is continuously weakened in the iterative process, and better prediction results are finally obtained.
\begin{figure*}[t]
    \vspace{-0.9cm}
    \setlength{\abovecaptionskip}{0.1cm} 
    \setlength{\belowcaptionskip}{0.1cm} 
    \centering
    \includegraphics[width=1.0\linewidth]{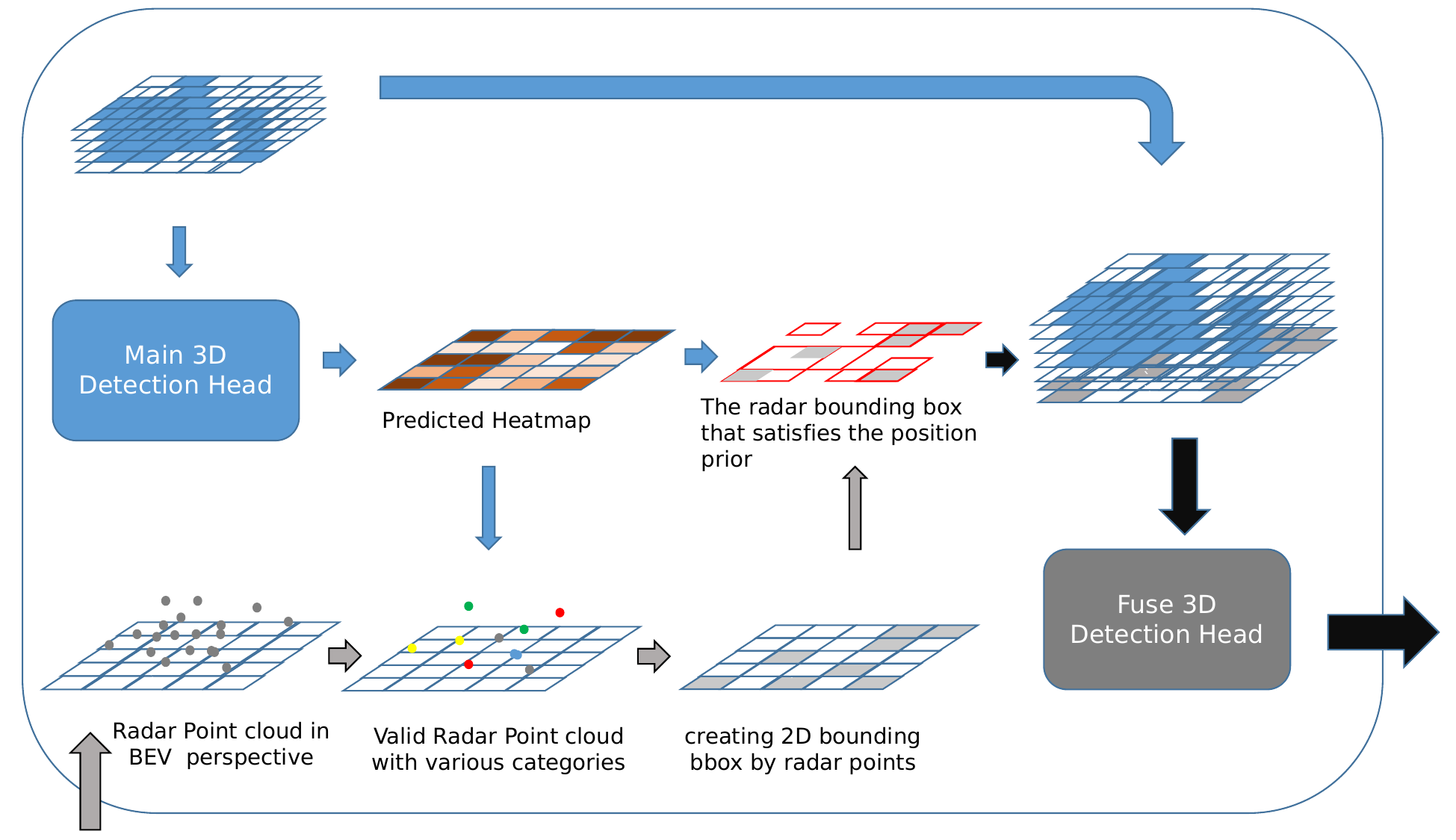}
    \caption{An image of clarify the process of radar data filtering and fusion}
    \label{fig:radar_process}
\end{figure*}
In the process of radar effective point matching, in order to improve the efficiency of matching, we adopted a two-stage matching method which includes the center point position matching and calculating the IOU matching scores between radar 2D bounding boxes and the ground truth. Before matching, the position of the ground truth under each category task may be redundant, we need to filter the ground truth to get a non-overlapping effective label bounding boxes. Then directly project the radar point cloud into the 2D grid which contains location information of the ground truth under BEV perspective, and record the projected points in label bounding boxes area as valid points, The part of point cloud which is in one-to-one correspondence with the filtered valid points will participates in the next stage matching process. Through the redundant filtering process with the label bounding boxes and the matching process of the projection points, the valid ground truth boxes area and the candidate radar point cloud with a small position offset can be efficiently picked on the 2D plane, effectively reducing the matching time and avoiding the impact on the matching accuracy when the size of the generated bounding boxes from radar points or the label bounding boxes generated by the main regression head are unstable.

In Figure \ref{fig:radar_process}, It shows the whole process of radar data processing, filtering, and fusion calculation. Radar data needs to be processed into point cloud format in the perspective of BEV, and then location prior information in heatmap which obtained by main 3D detection head with multiple categories filters the radar point cloud, then the obtained effective point cloud generates 2D bounding boxes in the BEV perspective,  which performs IOU matching with the reference valid area obtained by heatmap, finally the matched position information will be used to generate tensors related to radar position coordinates and velocity information, it will be fused with the original feature map and make predictions in the next step.

It should be noted that in the way of using the radar point cloud as an additional data set to improve the prediction effect, it is necessary to ensure that there is no deviation in the matching radar point cloud position. When we use the position predicted by the main regression head during the training process, the position prior which comes from predictions played an important role in the whole filtering process, so iterative training needs to be performed on the position prediction part of the main regression head in advance to ensure that the position prediction of the main regression head converges with the distribution of the final calibrated ground truth. At the same time, the stability of convergence at the final effect of the radar fusion method is guaranteed.
\section{Implementation Details}

\subsection{Training Settings}
For hyperparameters and backbones settings during our training, see Table \ref{tab:hyperparameters_training}. When a larger model is used or the BEV feature resolution is larger, the network tends to diverge, which can be alleviated to a certain extent by increasing the weight decay. In addition, if not specified, our experiments use 8 frames in the past and 8 frames in the future during training process. 

When using configuration 1 for training, we use the same backbone as Bevdet, and set the BEV size to 128*128, BEV channels is 80, this configuration is mainly used for small resolution image (e.g. 256*704) training and testing process, it is also used as the baseline-configuration during the entire optimization process. On this basis, we continue to optimize each stage of the training and testing process until more ideal evaluation scores is obtained. Configuration 2 is modified on the basis of configuration 1, it's used for the size adaptation experiment of backbone and BEV grid, and finally we propose configuration 3, which is used for the training and evaluation process of 640*1600 resolution images. At this time, larger parameters are chosen as BEV size and BEV channels. Fully retain the detailed information of large-resolution images to improve the accuracy of model prediction.
\begin{table}\scriptsize
  \centering
  \setlength{\tabcolsep}{0.5mm}
  \begin{tabular}{@{}lccccccc@{}}
    \hline
    Hyperparameter & Config 1 & Config 2 & Config 3 \\
    \hline
    backbone & Res50 & ConvNeXt-B or Intern-B & Intern-B  \\ 
    {epochs}          & 24         & 24         & 20           \\
    {batch size}      & 64         & 64         & 16           \\
    {optimizer}       & AdamW      & AdamW      & AdamW        \\
    {base lr}         & 0.0002     & 0.0002     & 0.0002       \\
    {weight decay}    & 0.01       & 0.05       & 0.05         \\
    {lr schedule}     & step decay & step decay & step decay   \\
    {warmup iters}    & 200        & 200        & 200          \\
    {warmup schedule} & linear     & linear     & linear       \\
    {gradient clip}   & 5          & 5          & 5            \\
    \hline
    {CBGS}            & \checkmark & \checkmark & \checkmark   \\
    {image size}      & 256×704    & 256×704    & 640x1600     \\
    {BEV size}        & 128        & 256        & 256          \\
    {BEV channels}    & 80         & 128        & 128          \\
    \hline
  \end{tabular}
  \caption{Training setting of HVDetFusion with different backbones}
  \label{tab:hyperparameters_training}
\end{table}

\subsection{Fusion Process}
We use the position regression value from main regression head in the HVDetFusion network trained for 6-8 epochs as the position prior to filter the false positive features of the radar point cloud. During this time, the position regression value is already very close to the true value in distribution. During the training and testing process, we use 51.2m in the physical dimension as the effective depth value to pick out the radar points that meet the distance requirements in the vehicle coordinate system. The number of rows and columns of the effective grid area in BEV perspective will be the same as  the resolution of the feature map which is calculated by the BEV encoder modules. The resolution is increased from 128*128 to 256*256 for higher prediction accuracy. The 2D bounding box generated by radar points uses 1m in the physical dimension as the reference distance of length and width. Due to the relatively poor accuracy of pedestrians and other objects during the collection process of radar datasets, we removed the fusion process of radar points for pedestrians and traffic cones. CenterHead is used as the regression head of the fusion feature, and we consider using the coordinates \(x\), coordinates \(y\), velocity components \(v_x\), and velocity components \(v_y\) in the radar datasets as effective information to generate multi-channel matrix as radar feature map, which will be fused together with the image feature map in a subsequent stage. In order to avoid that the distribution of radar point cloud in a single key frame is too sparse and cannot have a positive impact on the fusion effect, during the training process, the image in each key frame corresponds to the radar point cloud, which comes from the current frame, continuous previous two frames as supplements.
\section{Benchmark Results}
\label{sec:resu}
\subsection{Results on nuScenes val set}
\begin{table*}[h]\footnotesize
  \centering
  \vspace{0em} 
  \setlength{\tabcolsep}{1mm}
  \begin{tabular}{@{}lcccccccccc@{}}
    \hline
    Methods & Modality & Backbone      & Image Size & mAP↑  & NDS↑           & mATE↓ & mASE↓ & mAOE↓ & mAVE↓ & mAAE↓ \\ \hline
    {DETR3D}        & C        & ResNet50        & 900×1600   & 0.303 & 0.374          & 0.86  & 0.278 & 0.437 & 0.967 & 0.235 \\
    {BEVFormerV2}   & C        & ResNet-50       & 640×1600   & 0.349 & 0.428          & 0.75  & 0.276 & 0.424 & 0.817 & 0.193 \\
    {BEVDepth}      & C        & ResNet50        & 256×704    & 0.351 & 0.475          & 0.639 & 0.267 & 0.479 & 0.428 & 0.198 \\
    {BEVDet4D-Tiny} & C        & Swin-T          & 256×704    & 0.338 & 0.476          & 0.672 & 0.274 & 0.46  & 0.337 & 0.185 \\
    {SOLOFusion}    & C        & ResNet50        & 256×704    & 0.427 & 0.534          & 0.567 & 0.274 & 0.411 & 0.252 & 0.188 \\
    {VideoBEV}      & C        & ResNet50        & 256×704    & 0.422 & 0.535          & 0.564 & 0.276 & 0.44  & 0.286 & 0.198 \\
    {CRN}           & C,R      & ResNet-50       & 256×704    & \textbf{0.481} & \textbf{0.56}  & 0.474 & 0.271 & 0.541 & 0.328 & 0.188 \\
    \textbf{HVDetFusion} & C,R & ResNet-50   & 256×704    & \textbf{0.451} & \textbf{0.557} & \textbf{0.527} & \textbf{0.270} & \textbf{0.473} & \textbf{0.212} & \textbf{0.204} \\ 
    \hline
  \end{tabular}
  \caption{Comparison of different methods on the nuScenes val set.}
  \label{tab:contrast_net_score}
\end{table*}
In order to comprehensively compare the previous state-of-the-art 3D detection methods, we report the results on the nuScenes val  set in Table \ref{tab:contrast_net_score}. Besides the modality of camera-and-radar, the modality of camera-only is also listed. For fair comparison, we use ResNet-50 as the backbone of our networks which is used for comparing. Compared with other models, using a smaller resolution of 256×704, our method also exceeds most methods among camera-only and camera-and-radar modality, except that our method is 0.3\% lower than the NDS of the previous first-ranked CRN (HVDetFusion NDS 55.7\% vs CRN NDS 56\%). The data marked in bold in the last row of the table is the result of evaluation using HVDetFusion. Compared with most of the detection methods listed, we can obtain relatively higher mAP scores with small resolutions. This is the main factor for the improvement of NDS. In the displayed data of each object attribute, mAVE has been significantly reduced, which shows that using HVDetFusion can get more ideal results in predicting speed-related content, which may benefit from multiple frames fusion and the addition of radar data, the specific content will be shown in the ablation experiment.
\subsection{Results on nuScenes test set}
\begin{table*}[h]\footnotesize
\centering
\vspace{0em} 
\setlength{\tabcolsep}{1mm}
\begin{tabular}{@{}lcccccccccc@{}}
\hline
Methods                     & Modality & Backbone       & Image Size & mAP↑  & NDS↑           & mATE↓ & mASE↓ & mAOE↓ & mAVE↓ & mAAE↓ \\ 
\hline
CenterPoint-Single & L     & Voxel          & -        & 0.603          & 0.673          & 0.262  & 0.239 & 0.361          & 0.288  & 0.136 \\\hline
{FCOS3D}             & C   & R101-DCN       & 900×1600 & 0.358          & 0.428          & 0.69   & 0.249 & 0.452          & 1.434  & 0.124 \\
{DETR3D}             & C   & V2-99          & 900×1600 & 0.412          & 0.479          & 0.641  & 0.255 & 0.394          & 0.845  & 0.133 \\ 
{BEVFormer}          & C   & V2-99          & 900×1600 & 0.481          & 0.569          & 0.582  & 0.256 & 0.375          & 0.378  & 0.126 \\
{BEVDet4D}           & C   & Swin-B         & 900×1600 & 0.451          & 0.569          & 0.511  & \textbf{0.241}& 0.386  & 0.301  & 0.121 \\
{PETRv2}             & C   & Glom-like      & 640×1600 & 0.512          & 0.592          & 0.547  & 0.242 & 0.36           & 0.367  & 0.126 \\
{BEVDepth}           & C   & ConvNeXt-B     & 640×1600 & 0.52           & 0.609          & 0.445  & 0.243 & 0.352          & 0.347  & 0.127 \\
{SOLOFusion}         & C   & ConvNeXt-B     & 640×1600 & 0.54           & 0.619          & 0.453  & 0.257 & 0.376          & 0.276  & 0.148 \\
{VideoBEV-Base}      & C   & ConvNeXt-B     & 640×1600 & 0.554          & 0.629          & 0.457  & 0.249 & 0.381          & 0.266  & 0.132 \\
{BEVFormerV2}        & C   & InternImage-B  & 640×1600 & 0.54           & 0.62           & 0.488  & 0.251 & 0.335          & 0.302  & 0.122 \\
{BEVFormerOpt}       & C   & InternImage-XL & 640×1600 & 0.58           & 0.648          & 0.448  & 0.262 & 0.342          & 0.238  & 0.128 \\
{BEVDet-Gamma}       & C   & Swin-B         & 640×1600 & 0.586          & 0.664  & \textbf{0.375} & 0.243 & 0.377          & 0.174  & 0.123 \\
{VideoBEV}           & C   & ConvNeXt-B     & 640×1600 & 0.592          & 0.67           & 0.385  & 0.246 & \textbf{0.323} & 0.174  & 0.137 \\\hline
{CenterFusion}       & C,R & DLA34          & 450×800  & 0.326          & 0.449          & 0.631  & 0.261 & 0.516          & 0.614  & 0.115 \\
{CRAFT}              & C,R & DLA34          & 448×800  & 0.411          & 0.523          & 0.467  & 0.268 & 0.456          & 0.519  & \textbf{0.114} \\
CRN                  & C,R & ConvNeXt-B     & 640×1600 & 0.575          & 0.624          & 0.416  & 0.264 & 0.456          & 0.365  & 0.13  \\
\textbf{HVDetFusion} & C,R & InternImage-B  & 640×1600 & \textbf{0.609} & \textbf{0.674} & 0.379  & 0.243 & 0.382  & \textbf{0.172} & 0.132 \\
\hline
\end{tabular}
\caption{Comparison on the nuScenes test set}
\label{tab:score_on_test_set}
\end{table*}
\begin{table*}[h]
        \footnotesize
       \centering
       \vspace{0em} 
       \setlength{\tabcolsep}{1mm}
        \begin{tabular}{@{}l c c c c c c c c c c c c@{}}
           \hline
           & \multicolumn{2}{c}{Modality} & \multicolumn{10}{c}{mAP $\uparrow$} \\
           \cline{2-3} \cline{4-13}
           Method & C & R & Car & Truck & Bus & Trailer & Const. & Pedest. & Motor. & Bicycle & Traff. & Barrier\\
           \hline
           FCOS3D           & \checkmark & & 0.527 & 0.27   & 0.277 & 0.255  & 0.117 & 0.397  & 0.345 & 0.298 & 0.557 & 0.538 \\
           CenterNet (HGLS) & \checkmark & & 0.536 & 0.270  & 0.248 & 0.251  & 0.086 & 0.375  & 0.291 & 0.207 & 0.583 & 0.533 \\
           BEVDet           & \checkmark & & 0.643 & 0.35   & 0.358 & 0.354  & 0.162 & 0.411  & 0.448 & 0.296 & 0.601 & 0.614 \\
           BEVDet4D         & \checkmark & & 0.65  & 0.334  & 0.307 & 0.38   & 0.206 & 0.501  & 0.422 & 0.315 & 0.735 & 0.657 \\
           BEVDet-Gamma     & \checkmark & & 0.743 & 0.483  & 0.469 & \textbf{0.554} & 0.334 & 0.654  & 0.595 & 0.486 & 0.805 & 0.737 \\ \hline
           CenterFusion     & \checkmark & \checkmark & 0.509          & 0.258  & 0.234         & 0.235  & 0.077 & 0.370  & 0.314 & 0.201 & 0.575 & 0.484 \\
           CRN              & \checkmark & \checkmark & \textbf{0.784} & 0.526  & \textbf{0.57} & 0.483  & 0.292 & 0.582  & 0.634 & 0.478 & 0.749 & 0.647 \\
           HVDetFusion      & \checkmark & \checkmark & 0.761  & \textbf{0.531} & 0.527         & 0.513  & \textbf{0.338} & \textbf{0.668} & \textbf{0.651} & \textbf{0.51} & \textbf{0.827} & \textbf{0.768} \\ \hline
       \end{tabular}
       \caption{Per-class performance comparison for 3D object detection on nuScenes dataset.}
       \label{res:class_score}
   \end{table*}
    \begin{table*}[h]\scriptsize
    \centering
    \vspace{0em} 
    \setlength{\tabcolsep}{0.4mm}
    \begin{tabular}{lc@{\hskip0.1cm}c@{\hskip0.1cm}c@{\hskip0.1cm}c@{\hskip0.1cm}c@{\hskip0.1cm}c@{\hskip0.1cm}c@{\hskip0.2cm}c@{\hskip0.1cm}c@{\hskip0.1cm}c@{\hskip0.1cm}c@{\hskip0.1cm}c@{\hskip0.1cm}c@{\hskip0.1cm}c}
        \hline
    Method & IR & Backbone & PFN & FFN & Cam & Rad & PT & NDS $\uparrow$ & mAP $\uparrow$ & mATE $\downarrow$ & mASE $\downarrow$ & mAOE $\downarrow$ & mAVE $\downarrow$ & mAAE $\downarrow$ \\ 
        \hline
    Baseline & 256×704 & ResNet50   & 8 & 0 &  \checkmark & - & ImageNet & 0.533              & 0.419 & 0.584 & 0.272 & 0.419 & 0.269 & 0.219 \\
    Baseline & 256×704 & ResNet50   & 8 & 8 &  \checkmark & - & ImageNet & 0.557               & 0.451 & 0.527 & 0.270 & 0.473 & 0.212 & 0.204 \\ \hline
    Ours & 256×704  & ConvNeXt-B    & 8 & 8 &  \checkmark & - & ImageNet & 0.583               & 0.481 & 0.510 & 0.268 & 0.376 & 0.238 & 0.182  \\
    Ours & 256×704  & InternImage-B & 8 & 8 &  \checkmark & - & ImageNet & \textbf{0.587}      & 0.479 & 0.504 & 0.276 & 0.367 & 0.200 & 0.181 \\ \hline
    Ours & 256×704  & InternImage-B & 8 & 8 &  \checkmark & - & COCO     & \textbf{0.592}      & 0.487 & 0.501 & 0.269 & 0.367 & 0.199 & 0.181 \\
    Ours & 256×704  & InternImage-B & 8 & 8 &  \checkmark & - & OBJ365   & 0.589               & 0.485 & 0.501 & 0.278 & 0.359 & 0.220 & 0.180 \\ \hline
    Ours & 640×1600 & InternImage-B & 8 & 8 &  \checkmark & - & COCO     & \textbf{0.646}      & 0.562 & 0.453 & 0.251 & 0.309 & 0.187 & 0.153 \\ \hline
    Ours & 640×1600 & InternImage-B & 8 & 8 &  \checkmark & \checkmark & COCO & \textbf{0.652} & 0.571 & 0.450 & 0.243 & 0.316 & 0.174 & 0.150 \\ \hline
        \end{tabular}
         \caption{Overall ablation study on nuScenes validation set. (PFN: number of past frames, FFN: number of future frames, PT: use pretrained model, IR: Image Resolution)}
        \label{table:abl-score}
    \end{table*}
For the nuScenes test set, we train the HVDetFusion on the train and val sets. A single model with test time augmentation is adopted. We report the comparison results of the test set in Table \ref{tab:score_on_test_set}. HVDetFusion ranks first on the nuScenes camera-radar based 3D object detection leader board with a score of 67.4\% NDS, NDS is 5\% higher than the previous best method CRN, mAP is 3.4\% higher than CRN. The performance of our method outperforms most camera-only based methods such as FCOS3D, BEVFormerOpt, BEVDet-Gamma and VideoBEV. It also shows that as long as the radar is used properly, it is helpful to improve the perception effect of the camera.

Table \ref{res:class_score} further clarifies the performance effects of our proposed model on detection tasks with various categories. The bold mark in the table is the score with the best detection effect in each category. The data presented in this table are the test results published by the official list, where C represents the camera datasets, and R represents the radar datasets, the data in each column category represents the mean Average Precision in the corresponding detection results under the current category. The detection structure we proposed draws on the image processing method from the series of Bevdet network. After our optimization, the detection indicators of multiple categories have been improved by 3\%-6\%. Especially in the categories of Truck, Bus, and Motorcycle, the improvement is more obvious. Compared with the better-performing BevDet-Gamma detection network, the score increased by 4.7\% in the Truck category and 5.8\% in the Bus category.
The score of the Motorcycle category has increased by 5.6\%. Compared with the detection method that uses the camera dataset with the radar datasets at the same time, it has certain advantages in the detection effect of multiple categories. For example, compared with CRN, which has a better detection effect, the two methods are indistinguishable for categories with relatively large target sizes. However, in categories with relatively small spatial dimensions, our method has great advantages, especially Pedestrian , Traffic cone, and Barrier categories, our evaluation indicators are 8.6\%, 7.8\%, and 12.1\% higher than the corresponding categories. Through the comparison of each detection indicator in each category, we can clearly show the effectiveness of the method that our proposed.

\section{Ablation Study}
   In order to explore the influence of multiple experimental factors on the detection effect, we use the experimental method of controlling variables to conduct ablation experiments on the HVDetFusion network. The experimental results can refer to the overall evaluation indicators in Table \ref{table:abl-score}. During the ablation experiment, we uniformly use the verification datasets which provided by nuScenes officially as evaluation object. We use the HVDetFusion network with ResNet50 as the baseline to explore the influence of the backbone structure, the number of previous frames and future frames, whether to use the pre-trained model, the image resolution used and other factors on the detection results. 

   The part marked in bold in the table  \ref{table:abl-score} is the better indicator obtained during the detection process under the comparison of different factors. By comparing row 2 and row (3\&4) in the table, it can be found that there are differences in the detection results obtained by using different backbone structures, and better detection results can be obtained by using the ConvNeXt-Base or Internimage-Base structure. In our detection network, as shown in row (1\&2), we have achieved a substantial improvement by adding 8 future frames. Compared with the baseline, NDS has improved from 53.3\% to 55.7\%, which is an improvement of 2.4\%. Then we use the more advanced backbone, ConvNeXt-Base and Internimage-Base to do experiments. The ConvNeXt-Base network's NDS is increased by 2.6\%, and the Internimage-Base network's NDS is also slightly improved compared to the ConvNeXt-Base, with the NDS reaching 58.7\%. In the (5\&6) line, we compared and tested the effects of different pre-trained models. The COCO datasets pre-trained model has a better effect. Compared with the baseline which use Internimage-Base as backbone, NDS has improved 0.5\%. Then we tested the impact of the resolution on the model. It can be seen from lines (5\&7) that the 640x1600 resolution is 5.4\% higher than the 256x704 resolution on NDS indicator. Finally, combined with our fusion method, we introduce Radar Signal, NDS reached 65.2\%, which directly shows the effectiveness of radar fusion.

\section{Conclusion}

   In general, we propose a new detection algorithm called HVDetFusion, which is a multi-modal detection algorithm that not only supports pure camera data as input for detection, but also can perform fusion input of radar data and camera data. In the stage of processing camera data, we use a detection method based on Bevdet4D with structural optimization and improvement. This method can effectively extract the data of one or more camera sensors in key frames and integrate them into the BEV space, finally acquire good detection results. On this basis, we can consider integrating the radar sensor data at the same time, using the prior information of different object positions to filter the false positive information of the original radar data, and according to the positioning information and radial velocity information recorded by the radar sensors to supplement and fuse the BEV features generated by the original camera data, the effect is further improved in the process of fusion training.

\printbibliography

@inproceedings{li2022bevformer,
  title={Bevformer: Learning bird’s-eye-view representation from multi-camera images via spatiotemporal transformers},
  author={Li, Zhiqi and Wang, Wenhai and Li, Hongyang and Xie, Enze and Sima, Chonghao and Lu, Tong and Qiao, Yu and Dai, Jifeng},
  booktitle={European conference on computer vision},
  pages={1--18},
  year={2022},
  organization={Springer}
}

@inproceedings{caesar2020nuscenes,
  title={nuscenes: A multimodal dataset for autonomous driving},
  author={Caesar, Holger and Bankiti, Varun and Lang, Alex H and Vora, Sourabh and Liong, Venice Erin and Xu, Qiang and Krishnan, Anush and Pan, Yu and Baldan, Giancarlo and Beijbom, Oscar},
  booktitle={Proceedings of the IEEE/CVF conference on computer vision and pattern recognition},
  pages={11621--11631},
  year={2020}
}

@inproceedings{nabati2021centerfusion,
  title={Centerfusion: Center-based radar and camera fusion for 3d object detection},
  author={Nabati, Ramin and Qi, Hairong},
  booktitle={Proceedings of the IEEE/CVF Winter Conference on Applications of Computer Vision},
  pages={1527--1536},
  year={2021}
}

@article{huang2021bevdet,
  title={Bevdet: High-performance multi-camera 3d object detection in bird-eye-view},
  author={Huang, Junjie and Huang, Guan and Zhu, Zheng and Ye, Yun and Du, Dalong},
  journal={arXiv preprint arXiv:2112.11790},
  year={2021}
}

@article{huang2022bevdet4d,
  title={Bevdet4d: Exploit temporal cues in multi-camera 3d object detection},
  author={Huang, Junjie and Huang, Guan},
  journal={arXiv preprint arXiv:2203.17054},
  year={2022}
}

@article{liu2022petrv2,
  title={Petrv2: A unified framework for 3d perception from multi-camera images},
  author={Liu, Yingfei and Yan, Junjie and Jia, Fan and Li, Shuailin and Gao, Qi and Wang, Tiancai and Zhang, Xiangyu and Sun, Jian},
  journal={arXiv preprint arXiv:2206.01256},
  year={2022}
}

@misc{li2022bevdepth,
      title={BEVDepth: Acquisition of Reliable Depth for Multi-view 3D Object Detection}, 
      author={Yinhao Li and Zheng Ge and Guanyi Yu and Jinrong Yang and Zengran Wang and Yukang Shi and Jianjian Sun and Zeming Li},
      year={2022},
      eprint={2206.10092},
      archivePrefix={arXiv},
      primaryClass={cs.CV}
}

@misc{kim2022craft,
      title={CRAFT: Camera-Radar 3D Object Detection with Spatio-Contextual Fusion Transformer}, 
      author={Youngseok Kim and Sanmin Kim and Jun Won Choi and Dongsuk Kum},
      year={2022},
      eprint={2209.06535},
      archivePrefix={arXiv},
      primaryClass={cs.CV}
}

@inproceedings{wang2021fcos3d,
  title={Fcos3d: Fully convolutional one-stage monocular 3d object detection},
  author={Wang, Tai and Zhu, Xinge and Pang, Jiangmiao and Lin, Dahua},
  booktitle={Proceedings of the IEEE/CVF International Conference on Computer Vision},
  pages={913--922},
  year={2021}
}

@inproceedings{wang2022probabilistic,
  title={Probabilistic and geometric depth: Detecting objects in perspective},
  author={Wang, Tai and Xinge, ZHU and Pang, Jiangmiao and Lin, Dahua},
  booktitle={Conference on Robot Learning},
  pages={1475--1485},
  year={2022},
  organization={PMLR}
}

@inproceedings{park2021pseudo,
  title={Is pseudo-lidar needed for monocular 3d object detection?},
  author={Park, Dennis and Ambrus, Rares and Guizilini, Vitor and Li, Jie and Gaidon, Adrien},
  booktitle={Proceedings of the IEEE/CVF International Conference on Computer Vision},
  pages={3142--3152},
  year={2021}
}

@inproceedings{lin2020depth,
  title={Depth estimation from monocular images and sparse radar data},
  author={Lin, Juan-Ting and Dai, Dengxin and Van Gool, Luc},
  booktitle={2020 IEEE/RSJ International Conference on Intelligent Robots and Systems (IROS)},
  pages={10233--10240},
  year={2020},
  organization={IEEE}
}

@inproceedings{long2021radar,
  title={Radar-camera pixel depth association for depth completion},
  author={Long, Yunfei and Morris, Daniel and Liu, Xiaoming and Castro, Marcos and Chakravarty, Punarjay and Narayanan, Praveen},
  booktitle={Proceedings of the IEEE/CVF Conference on Computer Vision and Pattern Recognition},
  pages={12507--12516},
  year={2021}
}

@misc{liu2022bevfusion,
      title={BEVFusion: Multi-Task Multi-Sensor Fusion with Unified Bird's-Eye View Representation}, 
      author={Zhijian Liu and Haotian Tang and Alexander Amini and Xinyu Yang and Huizi Mao and Daniela Rus and Song Han},
      year={2022},
      eprint={2205.13542},
      archivePrefix={arXiv},
      primaryClass={cs.CV}
}

@article{li2022unifying,
  title={Unifying voxel-based representation with transformer for 3d object detection},
  author={Li, Yanwei and Chen, Yilun and Qi, Xiaojuan and Li, Zeming and Sun, Jian and Jia, Jiaya},
  journal={Advances in Neural Information Processing Systems},
  volume={35},
  pages={18442--18455},
  year={2022}
}

@article{kim2023crn,
  title={Crn: Camera radar net for accurate, robust, efficient 3d perception},
  author={Kim, Youngseok and Kim, Sanmin and Shin, Juyeb and Choi, Jun Won and Kum, Dongsuk},
  journal={arXiv preprint arXiv:2304.00670},
  year={2023}
}

@misc{pang2023transcar,
      title={TransCAR: Transformer-based Camera-And-Radar Fusion for 3D Object Detection}, 
      author={Su Pang and Daniel Morris and Hayder Radha},
      year={2023},
      eprint={2305.00397},
      archivePrefix={arXiv},
      primaryClass={cs.CV}
}

@inproceedings{yoo20203d,
  title={3d-cvf: Generating joint camera and lidar features using cross-view spatial feature fusion for 3d object detection},
  author={Yoo, Jin Hyeok and Kim, Yecheol and Kim, Jisong and Choi, Jun Won},
  booktitle={Computer Vision--ECCV 2020: 16th European Conference, Glasgow, UK, August 23--28, 2020, Proceedings, Part XXVII 16},
  pages={720--736},
  year={2020},
  organization={Springer}
}

@misc{yan2023cross,
      title={Cross Modal Transformer: Towards Fast and Robust 3D Object Detection}, 
      author={Junjie Yan and Yingfei Liu and Jianjian Sun and Fan Jia and Shuailin Li and Tiancai Wang and Xiangyu Zhang},
      year={2023},
      eprint={2301.01283},
      archivePrefix={arXiv},
      primaryClass={cs.CV}
}

@inproceedings{liu2022pai3d,
  title={Pai3d: Painting adaptive instance-prior for 3d object detection},
  author={Liu, Hao and Xu, Zhuoran and Wang, Dan and Zhang, Baofeng and Wang, Guan and Dong, Bo and Wen, Xin and Xu, Xinyu},
  booktitle={European Conference on Computer Vision},
  pages={459--475},
  year={2022},
  organization={Springer}
}

@article{xie2023sparsefusion,
  title={SparseFusion: Fusing Multi-Modal Sparse Representations for Multi-Sensor 3D Object Detection},
  author={Xie, Yichen and Xu, Chenfeng and Rakotosaona, Marie-Julie and Rim, Patrick and Tombari, Federico and Keutzer, Kurt and Tomizuka, Masayoshi and Zhan, Wei},
  journal={arXiv preprint arXiv:2304.14340},
  year={2023}
}

@misc{bai2022transfusion,
      title={TransFusion: Robust LiDAR-Camera Fusion for 3D Object Detection with Transformers}, 
      author={Xuyang Bai and Zeyu Hu and Xinge Zhu and Qingqiu Huang and Yilun Chen and Hongbo Fu and Chiew-Lan Tai},
      year={2022},
      eprint={2203.11496},
      archivePrefix={arXiv},
      primaryClass={cs.CV}
}

@misc{yang2022deepinteraction,
      title={DeepInteraction: 3D Object Detection via Modality Interaction}, 
      author={Zeyu Yang and Jiaqi Chen and Zhenwei Miao and Wei Li and Xiatian Zhu and Li Zhang},
      year={2022},
      eprint={2208.11112},
      archivePrefix={arXiv},
      primaryClass={cs.CV}
}

\end{document}